# Space and the Synchronic A-Ram.

## A General Purpose Language for a Massively Parallel, Formal Model of Computation.

Alex V Berka

Isynchronise Ltd.
email alex.berka@isynchronise.com

**Abstract**

Space is a new circuit oriented, spatial programming language designed to exploit the massive parallelism available in a novel formal model of computation called the Synchronic A-Ram, and physically related FPGA and reconfigurable architectures. Space expresses variable grained MIMD parallelism, is modular, strictly typed, and deterministic. Barring operations associated with memory allocation and compilation, modules cannot access global variables, and are referentially transparent. At a high level of abstraction, modules exhibit a small, sequential state transition system, aiding verification. Space deals with communication, scheduling, and resource contention issues in parallel computing, by resolving them explicitly in an incremental manner, module by module, whilst ascending the ladder of abstraction. Whilst the Synchronic A-Ram model was inspired by linguistic considerations, it is also put forward as a formal model for reconfigurable digital circuits. A programming environment has been developed, that incorporates a simulator and compiler that transform Space programs into Synchronic A-Ram machine code, consisting of only three bit-level instructions, and a marking instruction. Space and the Synchronic A-Ram point to novel routes out of the parallel computing crisis.

## 1. Introduction.

The pre-eminent issue facing computer science is the solution of conceptual and implementation problems relating to parallel computing. Multi-threading on the Von Neumann processor network remains arguably the leading theoretical approach for general-purpose parallelism, despite attempts to extend the applicability of SIMD/GPU model [1][2]. But multithreading has over four decades, struggled to enter a mainstream demanding the best available cost-performance ratios, because of what appear to be inherent problems in the model. Early doubts were expressed by proponents of Dataflow Models [3] concerning memory latency and synchronisation issues. A late nineties survey [4] observed that historically, the development of multi-threading and associated models was ad hoc, because they preceded the development of a general purpose, theoretical model of parallel computation. It proved to be difficult to devise multi-threading programming models that shielded the programmer from the tedious low level tasks of efficiently assigning threads and communication links to Von Neumann processing elements, and that provided a easy way for programmers to avoid resource contention [5][6][7][8].

### 1.1 Space Interlanguage and the Synchronic A-Ram.

The paper outlines a circuit oriented, spatial language environment for representing aspects of dataflow called *interlanguage*, which was used to derive a family of low level, finite and infinite formal models called the *α-Ram* family, including the finite *Synchronic A-Ram* [9]. The models currently lack a notion of propagation delay and are purely formal, to be simulated only. Their purpose is to help resolve conceptual issues in parallel environments and architectures. *Space* is a programming interlanguage for the Synchronic A-Ram, and may describe massively parallel algorithms at any level of abstraction, with the temporary exceptions of virtual functions and abstract data types. A programming environment called *Spatiale* [10] has been developed, and incorporates a simulator and compiler that transform Space programs into Synchronic A-Ram machine code, consisting of only four primitive instructions. Providing a programming methodology is adhered to, Space's runtime environment does not need to consider resource contention or deadlocks. Race and time hazards are resolved by local synchronisation. These features scale, and conceptually represent significant advantages over multithreading.

### 1.2 Sample module.

A toy Space module `bigaddition` is depicted in Figure 1, which employs an array of 65,536 sub-modules belonging to the `adder32` class, that adds two 32-bit integers with a maximum running time of 736 Synchronic A-Ram cycles. The code for `bigaddition` consists of numbered columnar representations called *interstrings* that express data movements and resource allocation relating to the preceding layer of abstraction, and sub-module activations. A control variable `i` is attached to parallel iteration *deep* constructs, which instruct the compiler to vertically replicate an interstring pattern 65,535 times.

```
module bigaddition{    // Line 1 loads the module's inputs into the adder array
                       // Line 2 activates adder array, and transfers adder outputs to module outputs
    storage{
        REG outputarray[65536] output;   // module has no inputs, only an output array
    };
    submodules{
      adder32 adder[65536];
    };
    replications{i/inc, 2*};      // declaration of control variable and incremental functions:
                                  // integer increment and multiply by 2
    time: 759-759 cycles;
    code{
       1.1:    #i -> adder[i].input0      :>  1: deep<i=0; i<= 65535; inc > (2,0) :;
              #i/2* -> adder[i].input1
       2.1: __adder[i] :: adder[i].output -> outputarray[i]  :>  2: deep<i=0; i<= 65535; inc > (3,0) :;
       3: HALT :;
    };
};
```

**Figure 1.** Space module that performs 65,536 distinct 32 bit integer additions.

The module seeds the adder32 submodule inputs with distinct immediate values determined by the control variable and an incremental function, and outputs the results into a register array. The module commences execution with the composite line 1. The deep structure mentions an *egress* line number (2,0), that transfers program control to line 2, upon completion of line 1. The second deep egress directs the module to a HALT. The module completes 65,536 simultaneous additions in 759 cycles. Spatiale compiles the module in half an hour, and simulation takes two and a half hours on a 1.5Ghz G4 processor. It is argued in [9] that it is neither feasible to simulate massive parallelism or high level computation on multi-tape Turing Machines, nor on graph rewriting versions of the λ-calculus, because of the models adverse complexity characteristics. Further parallel constructs in Space are discussed in section 5.

### 1.3 Organization of the paper.

Section 2 describes the background to interlanguages, which belong to the category of spatially oriented programming languages. Section 3 describes the Synchronic A-Ram formal model. Section 4 briefly overviews a language called *Earth*, which is close to the Synchronic A-Ram machine code, and allows the definition of the most primitive program modules used in Space. Section 5 presents Space, summarizing the type system and module declarations, and defines the basic interstring language structures, and presents program examples. Section 6 concludes by summarising the key benefits of Space and Synchronic A-Ram paradigm, henceforth termed *synchronic computation*, with respect to multi-threading and GPU computing.

### 2. Spatial Languages and Interlanguages.

Differing perspectives on what constitutes a spatially oriented programming approach were presented in [11] [12][13][14], centering around the notion of instructions being executed in situ without fetch, and program modules having a schematic or circuit character. The spatial approach questions the outlook that software may be studied as a pure discipline in isolation from hardware. For example, no pure λ-calculus interpreter and simulator for a high level language has been devised, that has acceptable complexity characteristics[9]. A spatial act of computation or communication is always linked in some way with a machine resource or channel. Verilog, VHDL, and configuration data for FPGAs and reconfigurable arrays of ALUs may be characterized as spatial. Interlanguages are generalized spatial languages, which provide a means of bypassing three structural features found in formal and programming languages, that have been carried over from human languages. It is argued the features have contributed towards preventing the emergence of a viable model of parallel computation.

### 2.1 Features associated with conventional tree syntax.

The programming of a novel parallel algorithm, in general requires an explicitly parallel language. The first feature relates to natural languages inability to express parallelism directly, so that many basic sentences may be conveyed simultaneously, thereby providing a cue for their meanings to be processed in parallel.

The second feature is called the Single Parent Restriction (SPR). SPR is the linguistic counterpart to the defining characteristic of trees; every node or part of speech may only participate in at most one more complex part of speech. SPR limits a part of speech describing an object, from participating directly in the expression of more than one relationship on a syntactic level, unless some form of naming, normally involving a semantic notion of storage, or sub-expression repetition is used. Repetition can result in an exponential increase in size for dataflow representations with respect to dataflow depth, compared with graph forms [9]. SPR is associated with trees with high structural variability, requiring complex parsing, and whose contents cannot easily be identified and accessed in parallel. Even if extensions are added to a tree language in order to describe parallelism on the syntactic level, as is the case with most parallel languages, SPR complicates the expression of shared structures in parallel processes.

The third feature relates to natural languages non-spatial character. They do not allow the tagging of abstract spatial information at the level of syntax, relating to data transfers and allocation of jobs to resources on a semantic level or preceding abstraction layer, that is argued in [9] is helpful in avoiding resource contention and state explosion in general purpose parallel computing. The claim that the three features are in fact serious defects relies on an argument that eliminating them from a language yields a better approach to parallel computation.

## 2.2 Related work dealing with language features.

Semantic networks [15] [16] and graph rewriting systems [17] [18] might be characterized as implicitly parallel formalisms, and addressed the SPR issue, but were non-spatial. They suffered from requiring pattern matching for an application of a deduction/rewriting rule, which needs the solution of the NP-complete sub-graph isomorphism problem. There are further combinatorial issues attendant upon implementing multi-threading on Von Neumann processor network implementations [9]. No semantic or graph machine models emerged.

The Dataflow Model (DM), developed in [18] [19] and [21] was another implicitly rather than explicitly parallel approach, that acknowledged the primacy of dataflow, and had a spatial character. Program control relied on a notion of a marking, consisting of a number of data tokens, being associated with the inputs of a functional node, being passed onto downstream functional nodes upon execution of a functional operation. Johnston et al in an influential survey [22], described software and some hardware related issues with DMs: the token system of program control cannot easily support iteration and data structures for dataflow programming languages, without adding significant complexity and potential deadlocks to programs. The latter placed severe limitations on DMs programmability. The survey also argued there was a mismatch between the fine-grained parallelism of dedicated dataflow architectures, and the coarse grained nature of many problems.

Finally, reference should be made to systolic programming models for coarse grained arrays of ALUs (CGAs) [12][14][23]. Systolic data sequencing and configuration software for CGAs bypass SPR, are spatial, and embody aspects of explicit parallelism, but lack a fundamentally novel theoretical model beyond the coarse grained systolic grid itself. Program control in systolic programming models relies on DM-type data tokens. The models are not general purpose, and suffer from being domain specific [24]. The developer is obliged to cast every program as a stream based program.

## 2.3 Dynamic semantics.

The counterargument that the language features do not need to be dealt with by syntax, because their effects can be handled by dynamic semantics is now briefly considered. Programmability issues, resource contention, and adverse cost-performance ratios associated with multithreading, have resulted in the *parallel computing crisis*. One may take the view that dynamic semantics is enough, and all is rosy in the garden of programming theory and conventional parallel architectures. But if interlanguage leads to a general-purpose compute model that circumvents programmability issues and resource contention, then it is likely that dynamic semantics only ever presented a partial solution, which discarded an opportunity to devise a better language structure.

## 2.4 Interlanguage = Interstring + Abstract Memory.

An interstring is a set-theoretical construct, designed for describing many-to-many relationships, dataflows, and simultaneous processes. It may be represented as a string of strings of symbol strings, where the innermost strings are short and have a maximum length. Interstring syntax is confined to a strictly limited range of tree forms, where only the rightmost, and the set of rightmost but one branches are indefinitely extendable. In conjunction with an abstract machine environment that does not reference semantics, an interstring can efficiently express at a syntactic level, sharing of subexpressions in a dataflow, scheduling, data transfers, and spatial allocation of machine resources, for the parallel processing of complex programs. Languages based on interstrings and abstract memories are called *interlanguages*, and were formally presented in [9]. The interlanguage environment bypasses the three natural language defects by sharing sub-expressions, by explicit parallelism, and by stipulating reference to data management and resources.

$$f(x,y,z) = (((x + y) + (y \times z) \times ((x + y) - (y \times z)))) \quad 1$$

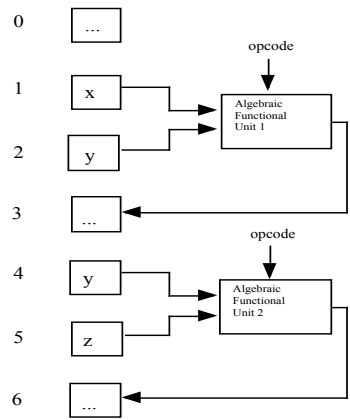

**Figure 2.** An abstract memory with abstract FU array.

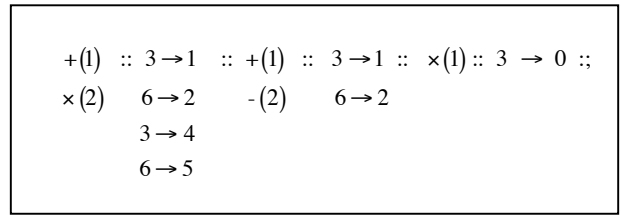

**Figure 3.** An interstring program which computes the function in equation 1, with the result stored in cell #0.

Figure 2 depicts a finite memory of 7 cells, each of which stores an element of the union of a set of variables and constants. The memory is attached to a notional array of 2 abstract functional units, but there may be any finite number of functional units with associated memory cells. Figure 3 depicts an interstring program, which evaluates the value of equation 1 in a synchronous parallel manner, if a semantic model is provided that interprets the meanings of the algebraic operators, and assigns values to the variables and constants from the semantic domain.

The interstring is composed of 6 columns of alternating types called *alpha* and *beta* columns, separated by double colons "::", and terminated by ";". The leftmost column is an example of an alpha column, which is a vertical string of pairs of function symbols and integers, represented as $f(j)$, which is understood to express the activation of the $j$th functional unit with the "opcode" $f$. The first and second inputs to the functional unit are the contents of the $3j+1$ and the $3j+2$ memory cells respectively. The result of applying the notional algebraic operation associated with the symbol $f$, is written into the $3j+3$ memory cell. The alpha column specifies the activation of all the mentioned functional units in one step, but there is a restriction that there is no more than one activation of the same functional unit.

A *beta column* is a string of pairs of integers, represented as $n\text{->}m$, each of which describes an instruction that copies "wirelessly" the contents of cell $n$, into cell $m$. The copies are understood to all occur in one step. Multiple reads from a cell are permissible, but multiple copies to the same cell are disallowed. It is also permissible for a cell to appear as both source and destination in a beta column, because a copy cycle consists of a read phase followed by a write phase. Given a semantic interpretation, an interstring applied to a memory, produces a sequence of memories, by executing an alternating succession of alpha and beta columns individually, from left to right. By convention, the result of evaluating the function $f$ described in equation 1, is copied into the zeroth cell by the final beta column.

A theorem is presented in [9], that asserts for any tree expression in the language of 2-ary functional terms in predicate logic, there exists a semantically equivalent interstring/memory pair. The theorem demonstrates that any arithmetic tree expression, which can exhibit high structural variability, can be coded by a simple finite memory, and an interstring whose tree structure has very restricted structural variability. Although the syntax and semantics for interlanguages are more complex than for functional terms in predicate logic, they do not exhibit the three defects described in section 2.

Alpha and beta columns express a synchronized form of parallelism, in which all operations in a column must cease before moving onto the next column. Upon initiating execution of the first interstring column, an interlanguage compiler may however duplicate the enhanced efficiency of the asynchronous, implicit parallelism found in DMs, where operations from differing dataflow columns may be triggered as soon as outputs from operations in earlier layers become available. The columnar format of interstrings then becomes a scheme for representing dataflows, rather than a dogmatic prescription for the scheduling of operations.

| Opcode | Destination cell ($x$) | Offset($y$) |
|---|---|---|
| bits 30-31 | bits 5-25 | bits 0-4 |

**Table 1.** Instruction format for 32-bit Synchronic A-Ram.

| Opcode | Instruction | Action |
|---|---|---|
| 00 | wrt0 $x\ y$ | write 0 into $(x,y)$ |
| 01 | wrt1 $x\ y$ | write 1 into $(x,y)$ |
| 10 | cond $x\ y$ | mark next register if $(x,y)=0$, else mark next but one register. |
| 11 | jump $x\ y$ | mark all instructions from register $x$, up to and including register $x+y$ |

**Table 2.** Instruction Set for Synchronic A-Ram.

## 3. The Synchronic A-Ram.

The development of the interlanguage environment gave pointers as to what might be needed from a purely formal model of parallel computation, which were mostly consistent with a spatially oriented outlook. Perhaps the most important insight was that any programmed act of computation should always be associated with an explicit location in which to store the result, which inspired the design of the α-Ram family of models [9], and the Synchronic A-Ram finite model in particular. The model not only provides a simple semantics for an interlanguage, but also suggests directions for developing novel architectures.

The Synchronic A-Ram is a globally clocked, fine grained, simultaneous read, exclusive write machine, and may be viewed as the lowest level, formal model for FPGAs. It incorporates a conventional, finite array of cells or registers composed of a constant number of bits, wherein the transmission of information between any two registers or bits occurs in one machine cycle.[1]

If the model's register's bit width is 32, there are 33,554,432 registers in the array. Let $(x,y)$ designate the $y$th bit of $x$th register, where $x$ is called the destination cell, and $y$ is called the offset. The Synchronic A-Ram's instruction format is displayed in Table 1, and the instruction set is in Table 2. Subject to some restrictions, any register is capable of either holding data, or of executing one of four primitive instructions in a cycle: the first two involve writing either '0' or '1' to $(x,y)$, the third instructs the register to inspect the bit stored in $(x,y)$, and select either the next or next but one register for activation in the following machine cycle, and the fourth is a jump which can activate the instruction in register $x$ in the following machine cycle, and also those in subsequent registers specified by the offset operand $y$.

### 3.1 Program control.

In the Von Neumann model, program control passes from one register containing an instruction to another, so that only one instruction is ever active per cycle. In the

---

[1] Although problematic from a physical standpoint, it is argued in [9] that this assumption can be worked around in deriving architectures, by incorporating a notion of signal propagation, and by the inclusion of Globally Asynchronous Locally Synchronous mechanisms for scaling.

Synchronic A-Ram model, multiple registers/instructions may be active per cycle, and execute in situ. Further, a Synchronic A-Ram assumes no ALU with the standard arithmetic functions; all operations have to be defined using only the four primitive instructions.

A *marking* represents a subset of registers, indicating which instructions (one per register) are to be executed in the next machine cycle. A machine state is the pair of the memory block and the marking, which initially only contains the instructions in registers 1 and 2. The Synchronic A-Ram's state transformation function is based on the instruction set, and an error detection scheme. A machine run consists of the state transformation function being repeatedly applied to the machine state either indefinitely, or until a special termination condition arises. An A-Ram process may therefore be seen as a sequence of pairs of memory blocks and markings. The model's formal definition in [9] incorporates error conditions that halt a machine run, if ill-defined or nonsensical behaviour is detected, including multi-set markings and multiple writes to the same bit. The removal of parallel-related contentions at the most primitive level of machine activity, contributes to eliminating contention in higher-level programming.

## 4. Earth.

Earth is a primitive parallel programming language, and is close to the level of Synchronic A-Ram machine code. Earth has numerous features [9], which facilitate the composition and readability of machine code. An Earth program is the most basic submodule that can be employed in a Space program. Earth and Space modules have a *level*, which is an unsigned integer representing the depth of module composition.

Earth is equipped with a *replicator* mechanism employing a control variable, for repeating code segments. In addition to VHDL and Verilog array type definitions of logic blocks and simple data transfer patterns between blocks, Earth replicators allow the concise definition of more complex logic gate and data transfer patterns.

Earth is powerful enough for modules to duplicate the functionality of complex sequential digital circuits, with a high degree of circuit parallelism. The size and depth of a sequential digital circuit, are approximately proportional to the size and cycle completion time respectively of an Earth implementation, providing the programmer matches the module's parallelism to the circuit's parallelism. Subject only to the constraint imposed by memory block size, the Earth programmer is perfectly at liberty to do this, given the highly parallel nature of the Synchronic A-Ram. Earth modules can readily implement primitive operations performed by devices such as *n*-input logic gates, register shifters, demultiplexers, incrementers, adders, and the arithmetic-logic functional units found in processor cores.

The code in figure 1(a) implements a sequential 4 input AND gate called *seqand4*, declares storage variables and their interface categories, much in the same style as a schematic VLSI module. Setting the *busy bit* indicates to the user that the machine is running. Resetting the busy bit indicates the machine has halted. An Earth *jump* instruction does not have to use an absolute register address *x*, and can employ a *relative* jump number, that refers to a labeling system in the code. The module also employs a replicator that generates the code in Figure 1(b). Note execution begins by marking the first two instructions.

```
NAME: seqand4;
BITS: busy private, output output;
BYTES: input input;   // leftmost 4 bits not used
TIME: 4-7 cycles; // min and max running times

        wrt1 busy
<0;i;3>{                // replicative structure
        cond input.i
        jump 1 1
}
        jump 3 1
1       wrt0 output     // first relative jump no.
        jump 2 0
2       wrt0 busy
3       wrt1 output
        jump 2 0
        endc            //  end of code
```

**Figure 4.** Replicative 4 bit AND gate.

```
        wrt1 busy
        cond input.0
        jump 1 1
        cond input.1
        jump 1 1
        cond input.2
        jump 1 1
        cond input.3
        jump 1 1
        jump 3 1
1       wrt0 output
        jump 2 0
2       wrt0 busy
3       wrt1 output
        jump 2 0
        endc
```

**Figure 5.** Replicated code for 4 bit AND gate.

## 5. Space.

Space was first presented in [9], and has a functionality comparable to C. Space is applicative, and is claimed to bypass the readability and efficiency issues associated with recursion based, functional style programming, whilst retaining verification friendly features such as state transition semantics, type strictness, and lack of side effects. The major features of Space are summarised in this section.

In common with an Earth module, a Space module is described in the spatial style as a hardware functional unit, and has names for the memory locations that hold the module's inputs, outputs, and internal storage. Space modules have declarations, one relating to pre-defined typed storage entities, and another to pre-defined library modules, called *submodules*. With some qualifications, a submodule does not retain a state between calls, or deliver differing outputs to identical inputs, making Space for the most part referentially transparent. In the role of a submodule, a module represents a category or class of processes, and has instances. A submodule cannot simply be called as an abstract software entity divorced from hardware. A submodule instance must be called, whose label is the first link in a chain, that reaches all the way down through layers of abstraction to machine resources.

```
module euclid{

    storage{
        unsigned a input;   // a must be greater than or equal to b
        unsigned b input;
        unsigned gcd output;
    };
    submodules{
        paror32 neqz; // 32 input OR gate functions as test for not equal to zero
        modulus mod; // modulus is based on somewhat slow subtraction based implementation
    };
    time: 1615-0 cycles;// min time is shown for a=b=1, max time is very large if a>>b=1

    code{

        1: b -> neqz.input    :: _neqz :: cond_neqz.output (3,0) (2,0) :;
            a -> mod.dividend
            b -> mod.divisor

        2: _mod :: mod.remainder -> neqz.input  :: _neqz :: cond_neqz.output (3,0) (2,0) :;
                mod.remainder -> mod.divisor
                mod.divisor -> mod.dividend

        3: mod.dividend -> gcd :: HALT :;   // transfer penultimate mod output to gcd
    };
};
```

**Figure 6.** A level two module implements Euclid's algorithm on two non-negative integers stored in *a* and *b*, where for simplicity the number stored in *a* is stipulated to be greater than or equal to the number in *b*. The greatest common divisor in *gcd* is obtained, by invoking the main loop described in line 2. The output *gcd* is recovered from the penultimate cycle's remainder, which continues to reside in mod.dividend. The code has no co-active parallelism. The only parallelism at the module's level of abstraction relates to data transfer in the copy columns of lines 1 and 2.

### 5.1 Type System and Storage Declarations.

Earth's types and some other basic types are pre-loaded into Space's type library. A new type may be derived by forming a construct whose members are pre-defined, existing types. The compiler calculates the space needed for a module's type declaration, by summing the amounts of registers that each member of the definition requires, based on its type and aggregate construct.

The four kinds of aggregate constructs are: singletons, arrays, pointers, and a linked list of arrays structure called a *blockstring*. A storage entity of a module of composition level *n*, has an *interface category*, which can indicate whether it is private to the module, and cannot be directly accessed by a module of level *n+1* or higher, or whether the entity is input or output or both, and is public, and therefore accessible to higher level modules, incorporating the module as a submodule.

The set of Space interface categories is identical to Earth's: *input, output, ioput,* and *private*. A storage declaration consists of a type name, a label representing an instance of the type, an aggregate construct (not needed for singletons), and an interface category. The type system is strict, in the sense that with a few exceptions involving low-level types, the contents of a storage entity, or of a storage entity associated with a submodule, can only be copied into the contents of another entity of the same type. At this stage of compiler bootstrapping, special modules must be written for each type, to implement memory allocation, pointer dereferencing, and access of array elements whose index varies at runtime. A block in memory is set aside from code to function as a heap during runtime.

### 5.2 Submodule Declarations.

The submodules declaration lists a series of member declarations of submodules, or arrays of submodules. A member declaration has a module class name, and a label name with aggregate construct, where the label represents a link to machine resources. A label name can assist the programmer in remembering any special role for the submodule(s). Labels facilitate software maintenance. If a submodule declaration is edited so that a sub-module class is substituted for an alternative, more efficient class with identical interface names, whilst retaining the same label, then no further editing of the module's code is required. Sub-modular aggregate constructs are restricted to singletons, and arrays with up to three dimensions.

### 5.3 Code.

Space's interstrings and program constructs enable the compact description of massively parallel code that incorporates resource allocation and data transfer management. Space code consists of a numbered sequence of interstrings called *base lines*, whose columns are separated by the '::' notation used in the interlanguage environment. A base line typically computes results of a dataflow, and the final column may also test a condition, and/or transfer control to other baseline(s). Some baselines may be attached by the notation ':>' to replicative and program control structures called *construct lines*. The Space compiler has a phase that processes and removes construct lines, leaving behind only (a possibly massive quantity of) base lines. A base line or construct line represents a subprogram of the module, with a single entry and at least one exit point. A construct line may have other construct-lines as components.

Both base and construct lines have a numerical label called a *line address*, which is a system where a string of integers is employed for the purpose of expressing component relationships between lines.

A base line is composed of a sequence of columns of instructions, drawn from a set of eight instructions, called the *base set*. The *copy* instruction employs an arrow, and transfers the contents of one storage entity to another of the same type, and corresponds to an interlanguage beta column. The *activation* instruction employs an underscore to indicate the activation of a submodule, and corresponds to an interlanguage alpha column. Activation of differing submodule classes in an activation column permits one kind of MIMD parallelism. A *cond* instruction tests a bit, and together with a *jump* can transfer program control to more than one base line by using an offset operand.

The deep construct has a single base line as it's only dependent line, and can express SIMD and limited SPMD parallelism. As exemplified in Figure 1, it defines a vertical replication of base-line code, in which a control variable is modified. In Figure 7, the *grow* construct is applied to a multi-line sub-program, and replicates the entire sub-program, in which a control variable is modified. Grow allows fully programmable SPMD parallelism within the module's level of abstraction. These constructs require barrier synchronisation in the current compiler implementation, and would be inefficient in a real environment with propagation delay. Alternative synchronisation mechanisms that allow local control whilst retaining global state transition semantics are under development.

### 5.3.1 Meta-modules.

Meta modules can retain a state by modifying a segment of their own code, and can then be separately instructed to execute that modified segment. Intentionally non-referentially transparent modules have roles in re-using code segments, and implementing high level programming features. A meta module's first phase is activated by the underscore, and performs the compiler-like task of modifying code. The second phase executes the modified code, and is activated using a hyphen.

### 5.3.2 Co-active Parallelism.

A deep-replicated baseline can only express a limited form of SPMD parallelism within a module, because there can be no explicit program control involving selection before the final column. In order to enable more complex forms of parallel programming, a second source of explicit parallelism in Space, is the ability to simultaneously activate multiple lines. A parallel algorithm often requires differing forms of replicated code to be active simultaneously. The presence of offsets in program control instructions in base-lines, and certain construct-lines, can instruct a number of subprograms represented by construct lines, and baselines to begin executing simultaneously, that will typically terminate at differing times. Such a set of lines is said to be *co-active*.

### 5.3.3 Containing Parallelism.

Unconstrained column and co-active parallelism have the potential to generate an undesirable number of states. A number of measures are taken to contain parallelism. Space does not allow the use of mixed base set instruction parallelism in a baseline column. Selection and jumps to other lines do not appear before the end of a base line, occurring only in the final base line column. Space is designed so that a module's behaviour may be characterized as a conventional sequential state transition system, where each state is associated with a set of co-active lines. The state system allows SIMD, M-SIMD, SPMD, MIMD, pipelining, systolic, cellular automata and other kinds of deterministic parallelism [9]. To achieve sequential state transition, a programming methodology is adopted, which constrains the way in which the programmer may invoke co-active parallelism.

A base line may not be activated, if it has not terminated from a previous activation. One base or construct line in a co-active set, is designated as the *carry line*, and takes as long or longer, to complete than the other lines. The carry line has the role of transferring control to the next state of the program (the next co-active set), at the end of its execution. The other members of the co-active set are forbidden from activating lines outside the co-active set, either whilst running, or upon their termination. The co-active sets that may be active at any stage of a module's run, are pre-determined at program composition time.

Scheduling, resource allocation and contention avoidance may be easily accomplished within the narrow confines of a module, and once resolved may be safely ignored at higher levels of abstraction.

## 6. Conclusion and Future Work.

The Synchronic A-Ram provides a simple semantics for exploring high-level deterministic parallelism in Space. If a notion of signal delay were to be introduced, the Synchronic A-Ram would represent a realistic abstraction of FPGA and CGA models. As a lower level model, it suggests novel mechanisms for general-purpose reconfigurable architectures, that are optimized for interlanguage program execution.

Interlanguage affords a means of bypassing the three defects that impede the description of parallelism, associated with formal and programming languages whose syntax has been templated from natural language. Interlanguages share subexpressions, support implicit as well as explicit parallelism, allocate resources, and facilitate the avoidance of resource contention.

The characterization of a Space module as a state transition system affords a means of avoiding deadlock and state explosion, making parallel programming almost as easy as sequential programming. Referential transparency, type strictness, and determinism will further assist the development of verification tools for Space, and contribute towards enhanced programmability compared with multithreading. Space is more general purpose than extensions of C and stream based programming models for reconfigurable platforms.

It is envisioned that synchronic computation will provide a synchronous, deterministic environment for general purpose parallel architectures, and largely leave asynchronous and non-deterministic features that improve efficiency to the compiler and runtime environment.

```
module addarray32{
 storage{
   unsigned A[32] input;
   unsigned sum output;
 };
 submodules{
     adder32 add[16];
     paror32 neqz;
     rightshift32 rightshift;   // register rightshift standing in for divide by two
     PJUMP{8} PJUMP;   // programmable jump, where offset can be varied during runtime.
     };
 replications{ i / inc, 2*, 2*+1};
 time: 0-0 cycles;
 code{
  1: jump (2,1) :;
  2: #8 -> PJUMP.offset       :: _PJUMP(5) :;  // sets PJUMP with first offset value
     #8 -> rightshift.ioput
3.1: A[i/2*]    -> add[i].input0  :: _add[i] :> 3: deep<i=0;i<=15; inc > (4,0)  :;
     A[i/2*+1] -> add[i].input1
  4: _rightshift :: rightshift.ioput -> PJUMP.offset :: _PJUMP(5) :; // activates PJUMP and main loop
     -PJUMP(5)       rightshift.ioput -> neqz.input       _neqz    // and then gives PJUMP new offset
5.1: add[i/2*].output -> add[i].input0   :: _add[i] :: jump(5.2,0) :> 5: grow<i=0;i<=7; inc > (6,0)  :;
     add[i/2*+1].output -> add[i].input1
5.2: subhalt(5) :;
  6: cond_neqz.output (7,0) (4,0) :;
  7: add[0].output -> sum :: HALT :;
 };
};
```

**Figure 7.** Parallel prefix adder for 32 integers. The PJUMP meta module can be programmed to vary jump offset during runtime. Used in conjunction with the grow construct, this facility reduces the number of instruction executions required to compute parallel prefix trees.

## References

[1] M. Harris, "Mapping computational concepts to GPUs," *ACM SIGGRAPH 2005 Courses*, Los Angeles, California: ACM, 2005, p. 50.

[2] J.D. Owens, D. Luebke, N. Govindaraju, M. Harris, J. Krüger, A.E. Lefohn, and T.J. Purcell, "A Survey of General-Purpose Computation on Graphics Hardware," *Computer Graphics Forum*, vol. 26, 2007, pp. 80-113.

[3] Arvind and R.A. Iannucci, "A critique of multiprocessing von Neumann style," *Proceedings of the 10th annual international symposium on Computer architecture*, Stockholm, Sweden: ACM, 1983, pp. 426-436.

[4] D.B. Skillicorn and D. Talia, "Models and languages for parallel computation," *ACM Comput. Surv.*, vol. 30, 1998, pp. 123-169.

[5] M. Flynn, "The Future Is Parallel But It May Not Be Easy," *High Performance Computing – HiPC 2007*, 2007, p. 1.

[6] K.E.A. Asanovic, *The landscape of parallel computing research: a view from Berkeley*, http://www.eecs.berkeley.edu/Pubs/TechRpts/2006/EECS-2006-183.html.: EECS at UC Berkeley,, 2006.

[7] E. Strohmaier, J.J. Dongarra, H.W. Meuer, and H.D. Simon, "Recent trends in the marketplace of high performance computing," *Parallel Comput.*, vol. 31, 2005, pp. 261-273.

[8] E.A. Lee, "The Problem with Threads," *Computer*, vol. 39, 2006, pp. 33-42.

[9] A.V. Berka, "Interlanguages and synchronic models of computation," *http://arxiv.org/pdf/1005.5183.*, May. 2010.

[10] http://sourceforge.net/projects/spatiale/

[11] F. Vahid, "It's Time to Stop Calling Circuits "Hardware"," *Computer*, vol. 40, 2007, pp. 106-108.

[12] R. Hartenstein, "The von Neumann Syndrome and the CS Education Dilemma," *Reconfigurable Computing: Architectures, Tools and Applications*, 2008, p. 3.

[13] M. Budiu, G. Venkataramani, T. Chelcea, and S.C. Goldstein, "Spatial computation," *SIGOPS Oper. Syst. Rev.*, vol. 38, 2004, pp. 14-26.

[14] A. DeHon, Y. Markovsky, E. Caspi, M. Chu, R. Huang, S. Perissakis, L. Pozzi, J. Yeh, and J. Wawrzynek, "Stream computations organized for reconfigurable execution," *Microprocessors and Microsystems*, vol. 30, Sep. 2006, pp. 334-354.

[15] W.A. Woods, *What's in a Link: Foundations for Semantic Networks,*, 1975.

[16] J.F. Sowa, "Principles of Semantic Networks," *Explorations in the Representation of Knowledge. Principles of Semantic Networks: Explorations in the Representation of Knowledge. Morgan Kaufmann*, 1991.

[17] M.R. Sleep, M.J. Plasmeijer, and M.C.J.D.V. Eekelen, Eds., *Term graph rewriting: theory and practice*, John Wiley and Sons Ltd., 1993.

[18] R. Plasmeijer and M.V. Eekelen, *Functional Programming and Parallel Graph Rewriting*, Addison-Wesley Longman Publishing Co., Inc., 1993.

[19] G. Kahn, "The Semantics of a Simple Language for Parallel Programming," *Information Processing '74: Proceedings of the IFIP Congress*, North-Holland, 1974, pp. 475, 471.

[20] Arvind and D.E. Culler, "Dataflow architectures," *Annual review of computer science vol. 1, 1986*, Annual Reviews Inc., 1986, pp. 225-253.

[21] J.B. Dennis and D.P. Misunas, "A preliminary architecture for a basic data-flow processor," *SCA*, 1975, pp. 126--132.

[22] W.M. Johnston, J.R.P. Hanna, and R.J. Millar, "Advances in dataflow programming languages," *ACM Comput. Surv.*, vol. 36, 2004, pp. 1-34.

[23] http://www.pact**xpp**.com/main/download/**XPP-III_programming**_WP.pdf

[24] W. Najjar and J. Villarreal, "Reconfigurable Computing in the New Age of Parallelism," *Embedded Computer Systems: Architectures, Modeling, and Simulation*, 2009, pp. 255-262.